%% file: main.tex
\pdfoutput=1
\documentclass[11pt]{article}

\usepackage{emnlp2021}

\usepackage{times}
\usepackage{latexsym}
\usepackage[T1]{fontenc}
\usepackage[utf8]{inputenc}
\usepackage{microtype}

\usepackage[ruled, linesnumbered, commentsnumbered, longend]{algorithm2e}
\usepackage{amsmath}
\usepackage{units}
\usepackage{graphicx}
\usepackage{multirow}

\input{mymacros.tex}

\title{Revisiting Tri-training of Dependency Parsers}

\author{Joachim Wagner \and Jennifer Foster \\
         School of Computing \\
         Dublin City University \\ Dublin, Ireland \\
         \texttt{firstname.lastname@dcu.ie}}

\begin{document}

\maketitle
\begin{abstract}
\input{002-abstract}

\end{abstract}

\section{Introduction}
\label{sec:intro}
\input{010-intro}

\section{Background}
\label{sec:previous}

\input{02x-previous/023-self}

\section{Experimental Setup}
\label{sec:setup}
\input{030-setup}

\section{Development Set Results}
\input{040-dev-results}

\section{Error Analysis}
\input{050-errors}

\input{072-table-test-results}

\section{Test Set Results}

\input{070-test-results}

\section{Conclusion}
\input{090-conclusion}

\section{Ethics and Broader Impact}

Tri-training uses much smaller amounts of unlabelled data than the
state-of-the-art semi-supervised method of self-supervised pre-training and we
therefore do not expect tri-training to add new risks from undesired biases
in the unlabelled data.
The use of tri-training may, however, pose new challenges in detecting
problematic effects of issues in unlabelled data as existing inspection
methods may not be applicable.

An individual tri-training run with FastText and multilingual BERT word embeddings
and $A=80k$, $T=8$ and $d=0.5$
typically takes three days on a single NVIDIA GeForce RTX 2080 Ti GPU.
Overall, we estimate that our experiments took 2500 GPU days.
This large GPU usage stems from the exploration of
tri-training parameters
in Appendix~\ref{sec:parameter-search}.
Future work can build on our observations and thereby reduce computational costs.

\section*{Acknowledgements}
\input{390-acknowledgements}

\bibliography{main}
\bibliographystyle{acl_natbib}

\appendix
\input{900-appendices}

\end{document}

%% file: mymacros.tex
\definecolor{DarkRed}{RGB}{140,4,0}
\definecolor{DarkGreen}{RGB}{0,102,5}
\definecolor{Grey}{RGB}{144,144,144}

\definecolor{ttblue}{RGB}{0,35,92}
\definecolor{ttred}{RGB}{90,9,0}
\definecolor{ttorange}{RGB}{121,89,0}
\definecolor{ttgreen}{RGB}{0,71,36}

\newcommand{\TODO}[1]{}
\newcommand{\NOTE}[1]{}

\newcommand{\ie}{i.\,e.\ }
\newcommand{\eg}{e.\,g.\ }

\newcommand{\etc}{etc.\ }

\newcommand{\udpf}{UDPipe-Future}

\DeclareMathOperator{\mmopsize}{size}

\DeclareMathOperator{\mmopreject}{reject\_duplicates}

%% file: 002-abstract.tex
We compare two orthogonal semi-supervised learning techniques,
namely tri-training and pretrained word embeddings,
in the task of dependency parsing.
We explore language-specific FastText and ELMo embeddings and
multilingual BERT embeddings.
We focus on a low resource scenario
as semi-supervised learning can be expected to have the most impact here.
Based on treebank size and available ELMo models, we select
Hungarian, Uyghur (a zero-shot language for mBERT) and Vietnamese.
Furthermore, we include English in a simulated low-resource
setting.
We find that pretrained word embeddings make more effective use of
unlabelled data than tri-training
but that the two approaches can be successfully
combined.

%% file: 010-intro.tex
Pre-trained neural architectures and contextualised word embeddings
are state-of-the-art approaches to combining labelled and unlabelled
data in natural language processing tasks taking
text as input.
A large corpus of unlabelled text is processed once and the resulting
model is either fine-tuned for a specific task or its hidden states are
used as input for a separate model.
In the task of dependency parsing, recent work is no exception to the
above.
However, earlier, pre-neural work explored many other ways to
use unlabelled data to enrich a parsing model.
Among these, self-, co- and tri-training had most impact~\cite{charniak-1997-statistical,steedman-etal-2003-bootstrapping,mcclosky-etal-2006-effective,mcclosky-etal-2006-reranking,sogaard-rishoj-2010-semi,sagae-2010-self}.

\emph{Self-training} augments the labelled training data with automatically
labelled parse trees predicted by a baseline model
in an iterative process:
\begin{enumerate}
\setlength\itemsep{0em}
    \item Select unlabelled sentences to be parsed in this iteration
    \item Parse sentences with current model
    \item Optionally discard some of the parse trees,
          \eg based on parser confidence
    \item Optionally oversample the original labelled data
          to give it more weight
    \item Train a new model on the concatenation of manually labelled
          and automatically labelled
          data
    \item Check a stopping criterion
\end{enumerate}
\emph{Co-training} proceeds similarly to self-training
but uses two different learners, each teaching the other learner,
\ie output of learner A is added to the training data of learner B
and vice versa.
\emph{Tri-training} uses three learners and only adds predictions to
a learner that the other two learners, the teachers,
agree on.
As with co-training, the roles of teachers are rotated so that
all three learners can receive newly labelled data.

We compare tri-training and
contextualised word embeddings in the task of dependency parsing,
using the same unlabelled data for both approaches.
In this comparison, we will try to answer:
\begin{enumerate}
\setlength\itemsep{0em}
    \item How does semi-supervised learning with tri-training compare
          to semi-supervised learning with a combination of
          context-independent and contextualised word embeddings?
    \item Are the above two approaches orthogonal, \ie do we get an
          additional boost if we combine them?
    \item How do these three approaches compare to the baseline of
          using only the manually labelled data?
\end{enumerate}

\noindent
We focus on low-resource languages as
\textit{(a)} maximising the benefits from semi-supervised learning even at
high computational costs is most needed for low-resource languages,
\eg to reducing editing effort in
the
manual annotation of additional
data, and
\textit{(b)} tri-training with high-resource languages comes at much
higher computational costs as not only is the manually labelled data
much larger but also the automatically labelled data can be expected
to need to be at least equally larger to have a relevant effect.
We select three low-resource languages,
namely Hungarian, Uyghur and Vietnamese,
(see Section~\ref{sec:languages} for selection criteria)
and,
English, simulating a low-resource scenario by sampling a subset of
the available data.

The results of our experiments show that 1) both tri-training and pretrained word embeddings offer an obvious improvement over a fully supervised approach, 2) pretrained word embeddings clearly outperform tri-training (between 2 and 5 LAS points, depending on the language), and 3) there
is some merit in combining the two approaches since the best performing model for each of the four languages is one in which tri-training is applied with models which use pretrained embeddings.

%% file: 02x-previous/023-self.tex
\subsection{Tri-training}

Tri-training has been used to tackle various natural language processing problems including
dependency parsing~\cite{sogaard-rishoj-2010-semi}, part-of-speech tagging~\cite{sogaard-2010-simple,ruder-plank-2018-strong}, chunking~\cite{chen-etal-2006-chinese}, authorship
attribution~\cite{qian-etal-2014-tri} and sentiment analysis~\cite{ruder-plank-2018-strong}. Approaches differ not only in the type
of task (sequence labelling, classification, structured prediction)
but also in the flavour of tri-training applied. These differences
take the form of the method used to introduce diversity into the three
learners, the number of tri-training iterations and whether
a stopping criterion is employed, the balance between manually and automatically labelled data, the selection criteria used to add an
automatically labelled instance to the training pool, and whether automatic labels from previous iterations are retained.

\newcite{zhou-li-2005-tri}
introduce tri-training.
They experiment with 12 binary classification
tasks with data sets from the UCI
machine learning repository, using bootstrap samples for model diversity.
Each pair of learners, the teachers,
sends their unanimous predictions to
the remaining third learner if
\textit{(a)} the error rate
             as measured on the subset of the
             manually labelled data which the two
             learners agree on
             is below a threshold
and
\textit{(b)} the total number of items that the teachers agree
             on and therefore can hand over to the learner
             reaches a minimum number that is adjusted in each round
             for each learner.
There also is an upper limit for the size of the new data
received that is enforced by down-sampling if exceeded. 
A learner's model is updated using the concatenation of the
full set of manually labelled data (before sampling) and the
predictions received from teachers.
If no predictions are received a learner's model is not updated.
Tri-training stops when no model is updated.

\newcite{chen-etal-2006-chinese}
apply tri-training to a sequence labelling task,
namely chunking, and discuss sentence-level
instance selection as a deviation from
vanilla tri-training.
They propose a ``two agree one disagree method''
in which the learner only accepts a prediction from its
teachers when it disagrees with the teachers.
\newcite{sogaard-2010-simple} re-invents this method
and coins the term
\emph{tri-training with disagreement} for it.

\newcite{li-zhou-2007-improve}
extend tri-training to more than three learners and
relax the requirement that all teachers must agree by
using their ensemble prediction.
They apply this to an ensemble of decision trees,
\ie a random forest, and call the method \emph{co-forest}.
As to the risk of deterioration of performance due to
wrong labelling decisions, they point to previous work
showing that the effect can be compensated with a sufficient
amount of data if certain conditions are met,
and they
include these conditions in the co-forest algorithm.

\newcite{guo-li-2012-improved}
identify issues with the update criterion
of tri-training and with the estimation
of error rates on training data and
propose two modified methods, one
improving performance in 19 of 33
test cases (eleven tasks and three learning
algorithms) and the other improving
performance in 29 of 33 cases.

\newcite{fazakis-etal-2016-self}
compare self-, co- and tri-training combined with a selection of
machine learning algorithms on 52 datasets and
include a setting where self-training is carried out
with logistic model trees, a type of decision tree classifier
that has logistic regression models at its leaf nodes.
Tri-training with C45 decision trees comes second in their
performance ranking after self-training with logistic model
trees.
However, logistic model trees are not tested with co- or tri-training.

\newcite{chen-etal-2018-tri}
adjust tri-training to neural networks by sharing parameters
between learners for efficiency.
Furthermore, they add random noise to the
automatic labels to encourage model diversity
and to regularize the models
and in addition to teacher agreement they
require teacher predictions made with
dropout (as in training) to be stable.

\newcite{ruder-plank-2018-strong} also
propose to share all but the final layers
of a neural model between the three learners
in tri-training for sentiment analysis and POS tagging.
They add an orthogonality constraint on the
features used by two of the three learners
to encourage diversity.
Furthermore,
they apply multi-task training in
tri-training
and
they modify the tri-training algorithm to
exclude the manually labelled
data from the training data of the third
learner.

\subsection{Tri-training in Dependency Parsing}

Tri-training was first applied in 
 dependency parsing by
\newcite{sogaard-rishoj-2010-semi},
who combine
tri-training with stacked learning in multilingual graph-based dependency parsing.
100k sentences per language are automatically labelled using three different stacks of token-level classifiers for arcs and labels, resulting in state-of-the-art performance on the CONLL-X Shared Task~\cite{buchholz-marsi-2006-conll}. 

In an uptraining scenario, \newcite{weiss-etal-2015-structured}
train a neural transition-based dependency parser
on the unanimous predictions of two slower, more accurate parsers.
This can be seen as tri-training with one
iteration and with just one learner's model
as the final model.
Similarly, \newcite{vinyals-etal-2015-grammar}
use single iteration, single direction tri-training in constituency parsing where the final model is a neural sequence-to-sequence model with attention, which learns linearised trees.

\subsection{Comparing Cross-view Training and Pretraining in NLP}

The only previous work we know of that compares
pretrained contextualised word embeddings to another
semi-supervised learning approach is the work of
\newcite{bhattacharjee-etal-2020-bert} who
compare three BERT models~\cite{devlin-etal-2019-bert} and a
semi-supervised learning method for neural models
in three NLP tasks:
detecting the target expressions of opinions,
named entity recognition (NER) and
slot labelling, \ie populating attributes of movies
given reviews of movies.
The semi-supervised learning method is
cross-view training \cite{clark-etal-2018-semi},
which adds auxiliary
tasks to a neural
network that are only given access to restricted
views of the input, similarly to the learners in
co-training, \eg a view may be the output of the
forward LSTM in a Bi-LSTM.
The auxiliary tasks are trained to agree with the
prediction of the main classifier on unlabelled
data.
Cross-view training performs best in NER and
slot labelling in 
\newcite{bhattacharjee-etal-2020-bert}'s experiments
and comes second and third place on two test sets
in opinion target expression detection.

%% file: 030-setup.tex
This section describes the technical details of
the experimental setup.

\subsection{Tri-Training Algorithm} \label{sec:setup:algo}
\input{03x-setup/031-description}

\subsection{Parameter Selection}
\label{sec:setup:parameters}
\input{03x-setup/032-parameters}

\subsection{Choice of Languages}
\label{sec:setup:languages}
\label{sec:languages}
\input{03x-setup/030-stats}

\input{03x-setup/033-languages}

\subsection{Unlabelled Data}
\label{sec:setup:unlabelled-data}
\input{03x-setup/034-unlabelled-data}

\subsection{Parser and Word Embeddings}
\label{sec:parser-and-embeddings}
\label{sec:setup:parser-and-embeddings}
\input{03x-setup/035-parser-and-embeddings}

\subsection{Ensemble Method for Candidate Models}
\label{sec:setup:ensemble}
\input{03x-setup/036-ensemble}

%% file: 03x-setup/031-description.tex
We provide an overview of our tri-training algorithm.\footnote{Full pseudocode
    is provided in Appendix~\ref{algorithm}.
    We share our source code, basic documentation and training log files
    (including development and test scores of each learner for all iterations)
    on \url{https://github.com/jowagner/mtb-tri-training}.
}
Before the first tri-training iteration,
three samples 
of the labelled data 
are
taken
and
initial models 
are trained on them.
Each tri-training iteration 
compiles three sets of automatically
labelled data,
one for each learner, 
feeding predictions that two learners agree on
to the third learner.\footnote{We require all
    predictions 
    (lemmata, universal and treebank-specific POS tags,
    morphological features,
    dependency heads and
    dependency labels including language-specific subtypes)
    for all tokens of a sentence to agree.
    The main reason is simplicity:
    The parser UDPipe-Future expects training data with all
    predictions as it jointly trains on them (multi-task learning).
    If we allowed disagreement between teachers on some of the tag
    columns we would have to come up with a heuristic to resolve such
    disagreements, complicating the experiment.
    Furthermore, we hypothesise that full agreement increases the
    likelihood of the syntactic prediction to be correct.
    The agreement can be seen as a confidence measure or quality
    filter.
}
In case all three learners agree, we randomly pick a receiving
learner.\footnote{Restricting  \label{fnt:choice3}
    the knowledge transfer to a single learner is
    a compromise between vanilla
          tri-training, which lets all three learners learn
    from unanimous predictions,
          and tri-training with disagreement \cite{chen-etal-2006-chinese}, which lets none of
    the learners learn from such predictions.
          Furthermore, this modification
          (together with rejecting duplicates while sampling 
          the unlabelled data 
          )
          increases diversity of the
          sets 
          and therefore may help
          keeping the learners' models diverse.
}
At the end of each tri-training iteration,
the three models 
are updated with new models trained on the concatenation of the
manually labelled and automatically labelled data selected for
the learners.

%% file: 03x-setup/032-parameters.tex
We explore three tri-training parameters:
\begin{itemize}
\setlength\itemsep{0em}
    \item $A$: the amount of automatically labelled data combined with labelled data when updating a model at the end of a tri-training iteration
    \item $T$: the number of tri-training iterations
    \item $d$: how much weight is given to
data from previous iterations.
The current iteration's data is always used in full.
No data from previous iterations is added with $d=0$.
For $d=1$, all available data is concatenated.
With $d<1$, we apply exponential decay to the dataset
weights, \eg for $d=0.5$ we take 50\% of the data
from the previous iteration, 25\% from the iteration
before the last one, \etc
\end{itemize}
For a fair comparison of tri-training with and without
word embeddings,
we take care that 
$A$, $T$ and $d$
are
explored equally well in both settings
and that each comparison is based on
results for the same set of
parameters.
Based
on the observations in
Appendices~\ref{sec:parameters:data-combination} to
\ref{sec:results:oversampling} and
balancing accuracy, number of runs
and computational costs,
we perform for each language and parser
twelve runs:
\begin{itemize}
\setlength\itemsep{0em}
    \item one run with $A=40$k, $T=12$ and $d=1$
    \item one run with $A=80$k, $T=8$ and $d=1$
    \item two runs with $A=80$k, $T=8$ and $d=0.5$
    \item two runs with $A=160$k, $T=4$ and $d=0.5$
    \item the above six runs in a variant where the seed data 
    is oversampled to match the size of 
    unlabelled data for the model updates at the end of each tri-training iteration.
\end{itemize}
For runs with multilingual BERT, we use $d \in \{0.5, 0.71\}$ instead of $d \in \{0.5, 1\}$
to reduce computational costs.

%% file: 03x-setup/030-stats.tex
\begin{table*}
\centering
\begin{tabular}{l|r|r|r|r|r|r|r|r}
 & \multicolumn{6}{c|}{\textbf{Labelled}} & \multicolumn{2}{c}{\textbf{Unlabelled}} \\
 & \multicolumn{2}{c|}{\textbf{Training}} & \multicolumn{2}{c|}{\textbf{Development}} & 
 \multicolumn{2}{c|}{\textbf{Test}} & 
 \multicolumn{2}{c}{\textbf{(filtered and sampled)}} \\
\textbf{Language} & \textbf{Tokens} & \textbf{Sent.}
 & \textbf{Tokens} & \textbf{Sent.}
 & \textbf{Tokens} & \textbf{Sent.}
 & \textbf{Tokens} & \textbf{Sentences} \\
\hline
English &
20,149 &
1,226 &
25,148 &
2,002 &
25,096 &
2,077 & 153,878,772 & 10,275,582 \\
Hungarian &
20,166 &
910 &
11,418 &
441 &
10,448 &
449 & 168,359,253 & 12,199,371 \\
Uyghur &
19,262 &
1,656 &
10,644 &
900 &
10,330 &
900 & 2,537,468 & 217,950 \\
Vietnamese  &
20,285 &
1,400 &
11,514 &
800 &
11,955 &
800 & 189,658,820 & 12,634,409 \\
\hline
\end{tabular}
\caption{Data statistics (UDPipe sentence splitting and tokenisation for the unlabelled data; right-most columns are for the unlabelled data used in tri-training; for
the ELMo and FastText training data, see Section~\ref{sec:setup:parser-and-embeddings})}
\label{tab:data:stats}
\end{table*}

%% file: 03x-setup/033-languages.tex
Since we focus on low-resource languages,
we select the three treebanks with the smallest amount of training
data from UD v2.3, meeting the following criteria:
\begin{itemize}
\setlength\itemsep{0em}

    \item The treebank has a development set.

    \item An ELMoForManyLangs model
          (Section~\ref{sec:parser-and-embeddings})
          is available for the target language.
          Sign languages and transcribed spoken treebanks
          are not covered.

    \item Surface tokens are included in the public UD release.

\end{itemize}
The treebanks selected are 
Hungarian \texttt{hu\_szeged} \cite{vincze-etal-2010-hungarian},
Uyghur \texttt{ug\_udt} \cite{eli-etal-2016-universal}, and
Vietnamese \texttt{vi\_vtb} \cite{nguyen-etal-2009-building}.
Furthermore, we include English in a simulated low-resource setting
using a sample of 1226 sentences (20149 tokens)
from the English Web Treebank \texttt{en\_ewt} \cite{silveira-etal-2014-gold}.
Table~\ref{tab:data:stats}
shows each treebank's training data size and
the
size of unlabelled data.
The sizes of the labelled training data
are in a narrow range of
19.3 to 20.3 thousand tokens.

%% file: 03x-setup/034-unlabelled-data.tex
To match the training data of the word embeddings
(Section~\ref{sec:parser-and-embeddings}), we use the Wikipedia and Common Crawl data
of the CoNLL 2017 Shared Task in UD Parsing
\cite{ginter-etal-2017-conll,zeman-etal-2017-conll}
as unlabelled data in tri-training.
We downsample the
Hungarian data to 12\%, the Vietnamese data to 6\% and the English
data to 2\% of sentences
to reduce disk storage requirements.
All data, including data for Uyghur, is further filtered
by
removing all sentences with less than five or more than
40 tokens\footnote{In preliminary experiments
    with the English LinEs treebank and without
    a length limit,
    learners rarely agree on predictions for
    longer sentences.
    This means that long sentences are unlikely to be
    selected by tri-training as new training data and
    the increased computational costs of parsing long
    sentences does not seem justified.
    We also exclude very short sentences as we do not
    expect them to feature new syntactic patterns and,
    if they do, to not provide enough context to infer
    the correct annotation.
}
and the order of sentences is randomised.
We then further sample the unlabelled data in each tri-training
iteration to a subset of fixed size to limit the parsing
and training costs.
The last two columns of Table~\ref{tab:data:stats} show
the size of the unlabelled data sets after filtering and
sampling.\footnote{These
    numbers do not reflect the removal of sentences
    that contain one or more tokens
    that have over 200 bytes in their UTF-8-encoded form
    and de-duplication performed before parsing unlabelled
    data.
}

%% file: 03x-setup/035-parser-and-embeddings.tex
For the parsing models of the individual learners in tri-training,
we use \udpf{} \cite{straka-2018-udpipe}.
This parser jointly predicts parse tree, lemmata,
universal and treebank-specific POS tags and
morphological features.
Since its input at predict time is just tokenised text, it can be
directly applied to
unlabelled data
while still exploiting lemmata and tags annotated in
the labelled data
to obtain strong models.
We use \udpf{} in two configurations:
\begin{itemize}
\setlength\itemsep{0em}
    \item \textbf{udpf}: \udpf{} with internal word and
          character embeddings only.
          This parser is for semi-supervised learning via
          tri-training only, \ie the unlabelled data only
          comes into play through tri-training.
          The parser's word embeddings are restricted to
          the labelled training data.

    \item \textbf{fasttext}: This parser is \udpf{} with
          FastText word embeddings \cite{bojanowski-etal-2017-enriching}.
          It is included to be able to tell how much of
          performance differences of the following two
          parsers is due to the inclusion of FastText word
          embeddings.

    \item \textbf{elmo}: This parser combines FastText word
          embeddings
          with ELMo
          \cite{peters-etal-2018-deep}
          contextualised word embeddings.
          This parser is for semi-supervised learning via
          training word embeddings on unlabelled data
          and via a combination of word embeddings and
          tri-training.

    \item \textbf{mbert}: This parser combines FastText
          word embeddings with multilingual
          BERT\footnote{\url{https://github.com/google-research/bert/blob/master/multilingual.md}}
          \cite{devlin-etal-2019-bert}
          contextualised word embeddings, pre-trained on
          Wikipedia in just over 100 languages including
          three of the four languages of our experiments.\footnote{To
              increase parser diversity in tri-training,
              each tri-training learner uses different
              BERT layers and subword unit vector pooling
              methods, see Appendix~\ref{sec:bert-layers}.
          }
          We include this parser to verify that our
          findings carry over to a transformer architecture.

\end{itemize}
\udpf{} employs external word representations without fine-tuning
the respective models, in our case FastText, ELMo and BERT.
Following \newcite{straka-etal-2019-evaluating}
          we use a fixed vocabulary of the one million
          most frequent types with FastText.
          We train FastText on the full CoNLL'17 data
          for each language separately, \ie
          9.4 billion tokens for English,
          1.6 billion tokens for Hungarian,
          3.0 million tokens for Uyghur and
          4.1 billion tokens for Vietnamese.
          FastText's feature to produce new
          word vectors for unseen words is not used.
          For ELMo, we use the 
          ELMoForManyLangs\footnote{\url{https://github.com/HIT-SCIR/ELMoForManyLangs}}
          models provided by \newcite{che-etal-2018-towards}.
          They limit training to ``20 million
          words'', \ie 0.2\% of the English data,
          1.2\% of the Hungarian data,
          100\% of the Uyghur data and
          0.5\% of the Vietnamese data.
Multilingual BERT is trained on Wikipedia only, presumably
on a newer data dump than the one used for the Wikipedia
part of the CoNLL'17 data.

An unusual feature of \udpf{} is that it oversamples the training data
to 9,600 sentences in each epoch if the training data is smaller than that.
This automatic oversampling enables the parser to perform well
on most UD treebanks without
tuning the number of
training epochs.
In our experiments, this behaviour will be triggered in settings with
a
low or medium augmentation size $A$, except when data of previous
iterations is combined ($d>0$) and the number of iterations $T$
is not small.\footnote{With $d>0$, the amount of training data
    grows with each iteration (up to a limit for $d<1$).
    For example, in the third tri-training iteration for Hungarian
    with mBERT, $d=0.71$ and $A = 40$k, learner 1 receives 3749 sentences
    (39986 tokens) from the current iteration, 2744 sentences from iteration 2,
    2038 sentences from iteration 1 and 2275 sentences from the labelled data
    (2.5 times 910 sentences), totalling in 10806 sentences which
    is above UDPipe-Future's oversampling threshold of 9600 sentences.
}

We make a small modification to the default learning rate schedule
of \udpf{}, softening its single large step from 0.001 to 0.0001
to five smaller steps,
keeping the initial and the final
learning rate.\footnote{We run \udpf{} with the option
    \texttt{--epochs 30:1e-3,5:6e-4,5:4e-4,5:3e-4,5:2e-4, 10:1e-4}
}
The seed for pseudo-random initialisation of the neural network of the
parser is derived from the seed that randomises the sampling of data in
each tri-training experiment, an identifier of the tri-training
parameters, an indicator whether the run is a repeat run,
the learner number $i$ and the tri-training
iteration $t$.

%% file: 03x-setup/036-ensemble.tex
Candidate models for the final model are created
at each tri-training iteration by
combining the current models of the three learners
in an ensemble using linear tree combination
\cite{attardi-dellorletta-2009-reverse}
as implemented\footnote{\url{https://github.com/jowagner/ud-combination}}
by
\newcite{barry-etal-2020-adapt}.\footnote{As we observed that performance of the ensembles on the
development data varies considerably
with the random initialisation of
the tie breaker in the combiner's greedy search, \eg{}
obtaining
LAS scores from 75.52 to 75.62 for ten repetitions, we run
the combiner 21 times with different initialisation
and report average LAS.}
Candidate ensembles are evaluated
on development data
using the CoNLL'18 evaluation script.

At tri-training iteration zero, \ie before any unlabelled data
is used, runs with the same language and parser
only differ in the random initialisation of
models.
A large number of additional ensembles can therefore be built,
picking a model for each learner from different runs.
These ensembles behave like ensembles from new runs,
allowing us to study the effect of random initialisation
using a much more accurate estimate of the LAS distribution
than possible with ensembles of each runs.
We obtain 4096 LAS values for each language and
parser as follows:
\begin{enumerate}
\setlength\itemsep{0em}
    \item For each learner $i \in \{1,2,3\}$,
          we partition the available models into 16
          buckets in order of development LAS.
    \item We enumerate all $16^3=4096$ combinations of
          buckets, and
          from each bucket combination, we sample one
          combinations of three models, one model per
          bucket.

    \item The selected model combination is combined using the linear tree combiner and  evaluated. 
\end{enumerate}

%% file: 040-dev-results.tex
\label{sec:results:main}
\input{04x-dev-results/041-main-comparison}

\input{04x-dev-results/041b-distribution}

%% file: 04x-dev-results/041-main-comparison.tex
We compare semi-supervised learning with
tri-training to semi-supervised learning with a combination
of context-independent and contextualised word embeddings,
and we compare these two approaches to combining them and
to training without unlabelled data, \ie supervised learning,
addressing the three questions posed in Section~\ref{sec:intro}.

As described in Section~\ref{sec:setup:parameters},
we obtain twelve LAS scores from tri-training with
each parser and language.
Tri-training with $T$ iterations can choose an ensemble
from $T+1$ ensembles (Appendix~\ref{sec:setup:model-selection}).
To give the baselines the same number of models to choose from,
we sample from the baseline ensembles described in
Section~\ref{sec:setup:ensemble}.
For each tri-training run with $T$ iterations,
we select the best score from $T+1$ baseline scores.
As the choice of the subset of $T+1$ scores is random,
we repeat the random sampling 250,000 times
to get an accurate estimate of the LAS distribution.
In other words,
    we simulate what would happen
    if the additional data obtained through tri-training
    had no effect on the parser.

Figure~\ref{fig:main-combined}
compares the parsing performance with and without
tri-training for the four development languages and
for the three types of parsing models \textbf{udpf}, \textbf{elmo} and \textbf{mBERT}.
Most distributions for each language are clearly separated
and the order of methods is the same:
both tri-training and external word embeddings yield
clear improvements.
External word embeddings have a much stronger effect
than tri-training.
In combination, the two
semi-supervised learning methods
yield a small additional improvement
with an average score difference of over half an LAS point.

%% file: 04x-dev-results/041b-distribution.tex
\begin{figure*}
  \centering
     \includegraphics[trim={0.8cm 11.8cm 0.75cm 0.55cm},clip,width=16cm]{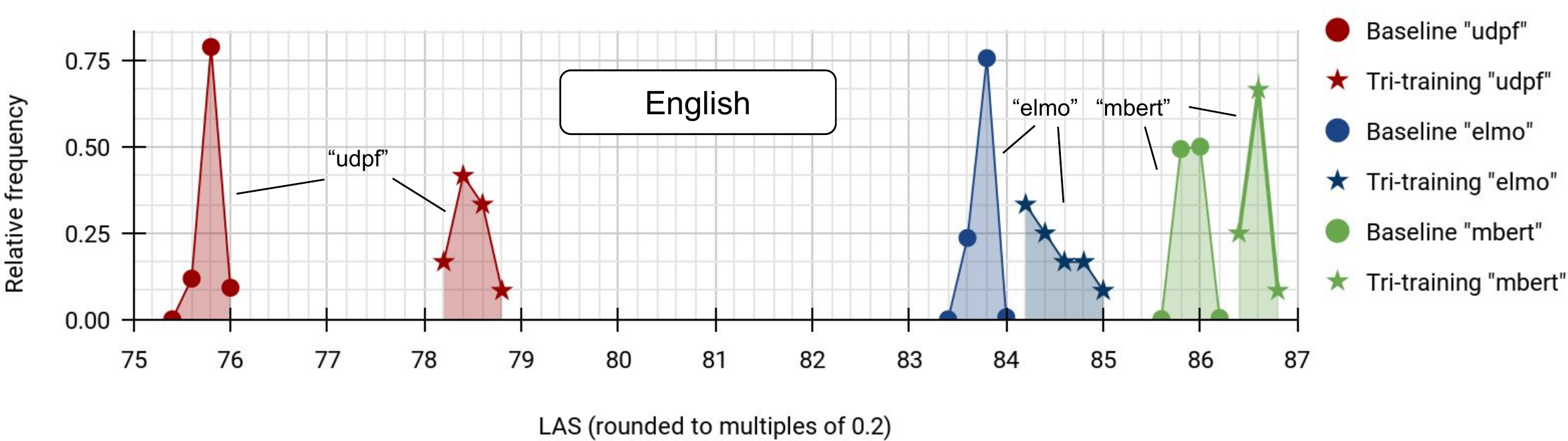}

  \vspace{0.3cm}

    \includegraphics[trim={0.8cm 11.9cm 0.75cm 0.55cm},clip,width=16cm]{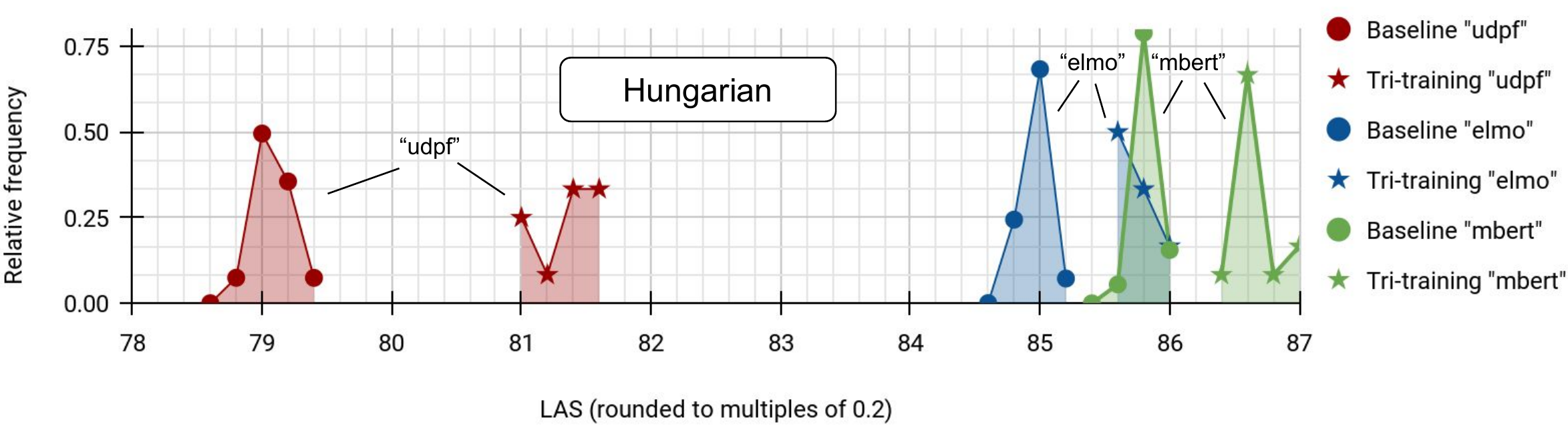}

  \vspace{0.3cm}

    \includegraphics[trim={0.8cm 11.9cm 0.75cm 0.55cm},clip,width=16cm]{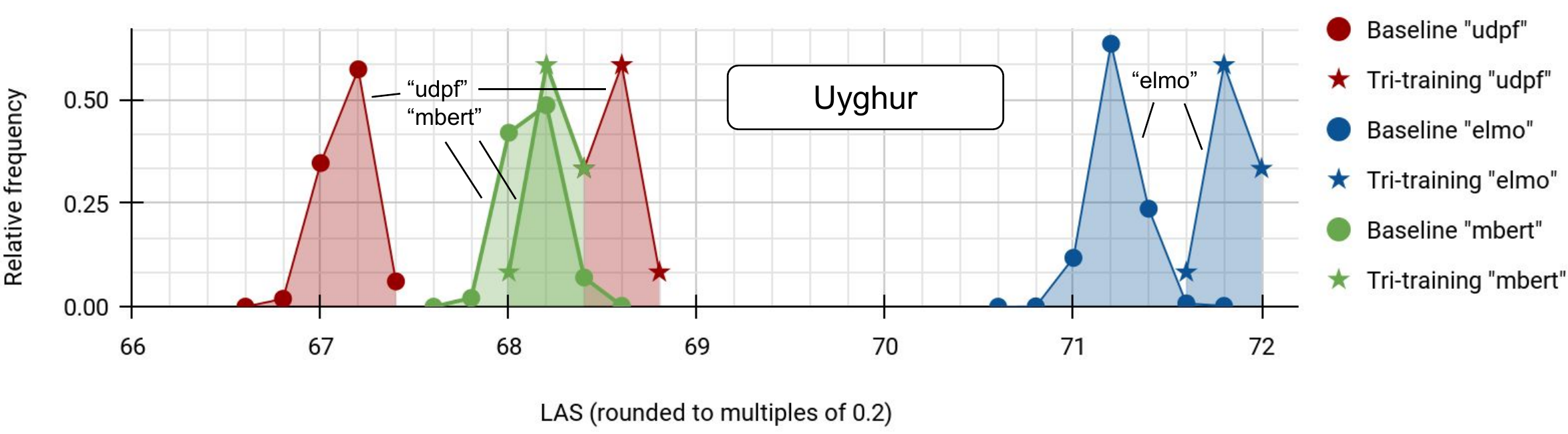}

  \vspace{0.3cm}

    \includegraphics[trim={0.8cm 11.9cm 0.75cm 0.55cm},clip,width=16cm]{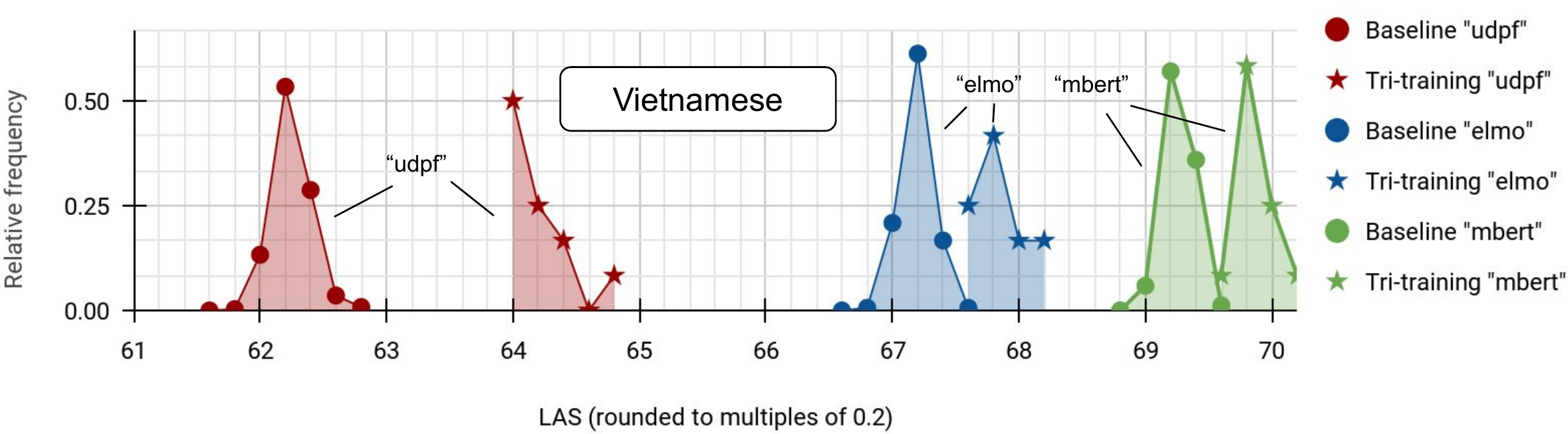}

  \vspace{0.3cm}

\caption{\label{fig:main-combined} Effects of external word embeddings and tri-training on the LAS distribution:
The baseline distributions are based on a large number of
ensemble LAS scores. Each tri-training distribution is based on twelve LAS scores.
}
\end{figure*}

%% file: 050-errors.tex
In this section, we probe, using the development sets, how the error
distribution changes as we add tri-training and/or word ELMo embeddings
trained on unlabelled data to the basic parser.
We compare the following
\begin{enumerate}
\setlength\itemsep{0em}
    \item  tokens that are out-of-vocabulary
relative to the manually labelled training data versus those that are in the training data (``OOV''/``IV'')
    \item different sentence length distributions: up to 9,
10 to 19,
20--39, and
40 or more tokens
\item different dependency labels, e.g is there a marked difference in the effect of tri-training or word embeddings for particular label types, e.g. \texttt{nsubj}?
\end{enumerate}

\paragraph{How does tri-training help (with no embeddings)?}
As expected, tri-training brings a clearly greater improvement for OOVs than IVs for all four languages. The role of sentence length is not consistent across languages. For English, tri-training helps most on longer sentences ($>$ 20 words), for Hungarian, short sentences ($<$ 10 words), and for Uyghur, 
 very long sentences ($>$ 40 words).
Sentence length does not appear to be a factor for Vietnamese. 
Regarding dependency labels, there are no 
clear pattern across languages.\footnote{See Table~\ref{table:LASbyLabel} in Appendix~\ref{error-dep-label}.}
\paragraph{How do word embeddings help (with no tri-training)?} 
Analysis of improvement by sentence length and by OOV status show similar trends to the tri-training improvements described above.
Across languages, the use of pretrained embeddings helps to correctly identify the \texttt{flat} relation (which is used in names and dates).

%% file: 072-table-test-results.tex
\begin{table*}
\centering
\begin{tabular}{l|l|l|l|l|l|l|l|l|}

 & \multicolumn{2}{c|}{\textbf{udpf}} & \multicolumn{2}{c|}{\textbf{fasttext}} & \multicolumn{2}{c|}{\textbf{elmo}} & \multicolumn{2}{c|}{\textbf{mbert}} \\
\textbf{Language} & \textbf{B} & \textbf{TT} & \textbf{B} & \textbf{TT} & \textbf{B} & \textbf{TT} & \textbf{B} & \textbf{TT} \\
\hline
\multicolumn{9}{l|}{Development} \\
\hline
English    & 75.8 & 78.8 & 78.1 & 80.6 & 83.7 &         84.9  & 85.9 & \textbf{86.7} \\
Hungarian  & 79.0 & 81.7 & 80.9 & 83.0 & 85.2 &         86.1  & 85.9 & \textbf{86.9} \\
Uyghur     & 67.3 & 68.8 & 68.4 & 69.7 & 71.2 & \textbf{72.1} & 68.1 &         68.4  \\
Vietnamese & 62.3 & 64.8 & 64.0 & 65.6 & 67.3 &         68.3  & 69.3 & \textbf{70.2} \\
\hline
\multicolumn{9}{l|}{Test} \\
\hline
English    & 76.6 & 79.3{$^{*****}$} & 78.9 & 80.8{$^{*****}$} & 84.0 &         84.9{$^{*****}$}  & 85.7 & \textbf{86.0}{$^{**}$} \\
Hungarian  & 77.6 & 79.2{$^{*****}$} & 79.8 & 81.2{$^{*****}$} & 84.3 &         84.9{$^{**}$}     & 85.5 & \textbf{86.3}{$^{***}$} \\
Uyghur     & 66.0 & 67.7{$^{*****}$} & 66.8 & 68.2{$^{*****}$} & 69.8 & \textbf{70.6}{$^{***}$}   & 67.4 &         67.5  \\
Vietnamese & 61.3 & 62.6{$^{****}$}  & 62.5 & 63.1{$^{*}$}     & 65.6 &         66.8{$^{****}$}   & 69.3 & \textbf{69.9}{$^{*}$} \\
\hline
\end{tabular}
\caption{Development and test set LAS for selected models (best of 12 according to development LAS); \textbf{B} = baseline, \textbf{TT} = tri-training; statistical significance of tri-training improvement over baseline (McNemar test; carried out for test set results only): one star for $p \leq 0.05$,
two stars for $p \leq 0.01$, three stars for $p \leq 0.001$, four stars for $p \leq 0.0001$ and
five stars for $p \leq 0.00001$.}
\label{tab:test-results}
\end{table*}

%% file: 070-test-results.tex
In this section, we verify to what extent our main observations
on development data carry over to test data and include results for
a parser using only FastText as external word embedding.
For each of the LAS distributions using tri-training
in
Figure~\ref{fig:main-combined}
and the distribution for \textbf{fasttext} not shown,
we select the ensemble with
highest development LAS for testing.
Since we also use model selection based on development LAS to choose
the final model of each tri-training run from its $T+1$ iterations,
the best model is selected from a set of 50 models given the values
of $T$ listed in Section~\ref{sec:setup:parameters}, exceeding the
number of baseline models available from iteration 0.
For a fair comparison, we therefore leverage the 4096 baseline ensembles
described in Section~\ref{sec:setup:ensemble}.
As a choice of 50 out of 4096 ensembles would introduce noise,
we repeatedly draw samples, for each sample find the best model
according to development LAS, obtain test LAS and report the average
LAS over all samples, \ie the expectation value.
As was the case for development results, we run the linear tree combiner
21 times on the three individual predictions of the tri-training
learners and take the average LAS over all combiner runs as the score
of the ensemble.

Table~\ref{tab:test-results}
shows development and test set LAS for the models
selected as described above.
The test set results confirm the development result
that a combination of
tri-training and contextualised word embeddings
consistently gives the best results and that
the individual methods improve performance.
In keeping with the development results, contextualised word embeddings
yield higher gains than tri-training.
The test results confirm the development observation that
multilingual BERT does not work as well as language-specific ELMo
for Uyghur, a zero-shot language for multilingual BERT.

%% file: 090-conclusion.tex
We compared two semi-supervised learning methods in the task of dependency parsing for three low-resource
languages and English in a simulated low-resource
setting.
Tri-training was effective but could not come close
to the performance gains of contextualised word embeddings.
Combined, the two learning methods achieved small
additional improvements between 0.2 LAS points for Uyghur
and 1.3 LAS points for Vietnamese.
Whether these
gains can justify the additional
costs of tri-training will depend on the application.

We recommend that users of tri-training vary
settings and repeat runs to find good models.
Future work could therefore explore how to
best combine the many models or the large
amount of automatically labelled data that such
experiments produce.
To obtain a fast and strong final model,
a combination of ensemble search and model distillation or up-training may be the next
step.
Integrating cross-view training \cite{clark-etal-2018-semi}
into tri-training may also be fruitful
similarly to the integration of multi-view learning
in co-training \cite{lim-etal-2020-semi}.
The requirement of tri-training
that two teachers must agree
changes the sentence length distribution of
the data selected and may introduce other
biases.
Future work could try to counter this effect
be re-sampling the predictions similarly to
how \newcite{droganova-etal-2018-mind} corrected
for such effects in self-training.

While our literature review suggests that tri-training performs better
than co- and self-training, it would be interesting how these methods
compare under a fixed computation budget as the latter methods train
fewer parsing models per iteration.

%% file: 390-acknowledgements.tex
This research is supported by Science Foundation Ireland (SFI)
through the ADAPT Centre for Digital Content Technology, which is
funded under the SFI Research Centres Programme (Grant 13/RC/2106)
and is co-funded under the European Regional Development Fund, 
and through the SFI Frontiers for the Future programme (19/FFP/6942).

%% file: 900-appendices.tex
\section{Tri-training Algorithm}\label{algorithm}
\input{90x-appendices/0311-algorithms}
\input{90x-appendices/0312-full-description}

\section{Parameter Search}\label{sec:parameter-search}

\input{90x-appendices/0600-parameter-search}

\section{Error Analysis: Dependency Labels}\label{error-dep-label}
\input{90x-appendices/0760-error-analysis}

\section{Learning Rate Schedule}

\input{90x-appendices/0730-learning-rate-scehd}

\section{Model Selection} \label{sec:setup:model-selection}
\input{90x-appendices/0380-model-selection}

\section{BERT Layer Selection}\label{sec:bert-layers}
\input{90x-appendices/0750-bert-layers}

%% file: 90x-appendices/0311-algorithms.tex
\begin{algorithm}
\caption{Tri-training in this work.
}
\label{algo:tt}

\KwIn{$L$: Labelled data \newline
      $U$: Unlabelled data \newline
      $A$: Maximum number of items to add per iteration and learner \newline
      $T$: Number of tri-training iterations \newline
      $d$: Decay parameter, $0 \leq d \leq 1$
      }
\KwOut{Models $\{h_1, h_2, h_3\}$ for ensemble.}
\For{$i \in \{1,2,3\}$}{
    $B_i \leftarrow$ Sample$(L, \hspace{0.2em} \mmopsize = 2.5 \times |L|)$

    $h_i \leftarrow$ Learn$(B_i)$
}
\For{$t=1$ to $T$}{
    $U' \leftarrow$ Sample$(U, \hspace{0.2em} \mmopsize{} = 16 \times A), $\\
    \hspace{2.5em} $\mmopreject{} = $ \emph{True}$)$
    \label{algo:tt:subset}

    $L_{t,i} \leftarrow \{\}, \quad i=1,2,3$

    \For{$x \in U'$}{
        \If{$h_j(x) = h_k(x), j \neq k$}{  \label{algo:tt:hjhk}
            \eIf{$h_1(x) = h_2(x) = h_3(x)$}{
                 $C \leftarrow \{1,2,3\}$
            }{
                 $C \leftarrow \{1,2,3\} \setminus \{j,k\}$
            }
            $i \leftarrow $RandomChoice$(C)$

            $L_{t,i} \leftarrow L_{t,i} \cup \{(x, h_j(x))\}$
            \label{algo:tt:xhjx}
        }
    }
    \For{$i \in \{1,2,3\}$}{
        \If{$|L_{t,i}| > A$}{
            $L_{t,i} \leftarrow$ Sample$(L_{t,i}, \hspace{0.2em} \mmopsize{} = A)$
            \label{algo:tt:limit-lti}
        }

        \label{algo:tt:downto}
        $R \leftarrow \bigcup\limits_{t'=1}^{t} $ Sample$(L_{t',i}, \mmopsize{} =$ \label{algo:tt:new:start} \\
        \hspace{4.1em} $ \min \{|L_{t',i}|, {A \times d^{t-t'}}\})$
            \label{algo:tt:new:end}
        \vspace{0.5em}

        $h_i \leftarrow$ Learn$(B_i \cup R)$
        \label{algo:tt:update} 
    }

}
\end{algorithm}

%% file: 90x-appendices/0312-full-description.tex
Algorithm~\ref{algo:tt}
shows the tri-training algorithm in the form we use it in
this work.
An extended version of the description in Seciton~\ref{sec:setup:algo} follows.

\paragraph{Lines 1--3}

Before the first tri-training iteration, three samples $B_i$
of the labelled data $L$ are taken and initial models $h_i$
are trained on them ($i \in \{1,2,3\})$.
We sample without replacement and with a target size 2.5 times
the size of $L$, re-populating the sampling urn
each time it becomes empty.\footnote{\newcite{zhou-li-2005-tri}
    sample with replacement.
    Section~\ref{sec:parameter:seed-sampling} motivates
    our choice.
}

\paragraph{Lines 5--7}

For each tri-training iteration, we further sample a
de-duplicated subset $U'$
of the unlabelled data as processing all available
unlabelled data would not be practical for most
languages in our experiments
(Section~\ref{sec:setup:unlabelled-data}).\footnote{In \label{fnt:sampling-u}
    \newcite{zhou-li-2005-tri}'s experiments, all
    datasets are small, that largest having 3772 items.
 }$^{,}$\,
\footnote{We set the
    size of $U'$ so that we do not expect it to
    be a limiting factor.
    In preliminary experiments with the English LinEs
    treebank, we observed that $4A$ is sufficient to
    obtain at least $A$ new labelled items.
    We set the size of $U'$ to $16A$ to
    account for likely variation in the rate of
    agreement between models when switching to other
    treebanks.
}

\paragraph{Lines 5, 8--18}

Each tri-training iteration 
$t$ 
compiles three sets of automatically
labelled data
$L_{t,i}$
one for each learner $i$,
feeding predictions that two learners agree on\footnote{See Footnote 2.}
to the third learner
(lines~\ref{algo:tt:hjhk}--\ref{algo:tt:xhjx}).
In case all three learners agree, we randomly pick a receiving
learner.\footnote{See Footnote~\ref{fnt:choice3}.}
While
\newcite{zhou-li-2005-tri}
do not state in their pseudo code that $j$ and $k$ must be different,
this is clear from their description.

\paragraph{Lines 21 and \ref{algo:tt:limit-lti}}

We limit the size of the data sets $L_{t,i}$ to $A$, downsampling
them if needed.

\paragraph{Lines 19, 23--35}

While \newcite{zhou-li-2005-tri} update the models $h_i$
by directly
training on $L \cup L_{t,i}$, we experiment with
concatenating data from previous tri-training
iterations (lines~\ref{algo:tt:new:start}--\ref{algo:tt:new:end})
and we use $B_i$ 
instead of
$L$ (line~\ref{algo:tt:update}).
\footnote{Initially,
     the latter was an error on our side but, given
          that we ensure that each item in $L$ is
          included in each $B_i$ at least twice,
          keeping the $B_i$ seems better fitted as $L$ no
          longer provides additional labelled
          data, the diversity of learner models is improved
          and moderate oversampling of $L$ can be expected to
          be helpful.
}
The parameter $d$ controls how much weight is given to
data from previous iterations.
The current iteration's data is always used in full.
No data from previous iterations is added with $d=0$.
For $d=1$, all available data is concatenated.
With $d<1$, we apply exponential decay to the dataset
weights, \eg for $d=0.5$ we take 50\% of the data
from the previous iteration, 25\% from the iteration
before the last one \etc
(line~\ref{algo:tt:new:end}).
\footnote{We
    do not restrict experiments to a single value of $d$
    as tri-training is considerably faster with $d \in \{0, 0.5\}$
    than for $d=1$, see
    Section~\ref{sec:parameters:data-combination}.
}
At the end of each tri-training iteration 
 in
(line~\ref{algo:tt:update}),
the three models 
$h_i$
are updated with new models trained on the concatenation of the
manually labelled and automatically labelled data selected for
the learners.\footnote{For
    \label{fnt:update-criteria}
    the reasons described in
    Section~\ref{sec:setup:model-selection},
    we do not use the model update conditions
    of \newcite{zhou-li-2005-tri}
    that are based on
    \textit{(a)}
          the estimated label noise in $R$ to be lower than
          in the previous iteration and on
    \textit{(b)}
          $R$ to be
          sufficiently big for the noise not to be harmful
          under certain assumptions. 
}

Our training.conllu files produced by tri-training for each
parsing model start with the manually labelled data
followed by the automatically labelled data.
In case of oversampling $B_i$ to match the size of $R$
(option not mentioned in the description above),
the oversampling also changes the order of the manually
labelled data.
Similarly, the $L_{t',i}$ are only re-ordered if
$|L_{t',i}| > A \times d^{t-t'}$.

For clarity, when we use the set union operator in
Algorithm~\ref{algo:tt} we mean concatenation of data sets.
Duplicates are not removed.
It is also clear that vanilla tri-training concatenates
sets as set operations in the mathematical sense would
damage the samples with replacement of manually labelled
data.

%% file: 90x-appendices/0600-parameter-search.tex
This section describes our parameter search,
analysing four distinct aspects of tri-training:
sampling of seed data, reusing data from previous iterations,
sample size and oversampling.

\subsection{Effect of Sampling of Seed Data}
\label{sec:parameter:seed-sampling}
\label{sec:results:seed-sampling}
\input{90x-appendices/0610-seed-sampling}

\subsection{Effect of Using Data from Previous Iterations}
\label{sec:parameters:data-combination}
\label{sec:results:data-combination}
\input{90x-appendices/0620-data-combination}

\subsection{Effect of Sample Size $A$}
\label{sec:results:augsize}
\input{90x-appendices/0630-augment-size}

\subsection{Effect of Oversampling}
\label{sec:results:oversampling}
\input{90x-appendices/0640-oversampling}

%% file: 90x-appendices/0610-seed-sampling.tex
\subsubsection{Seed Sampling Methods Considered}
\label{sec:seed-data}
\input{90x-appendices/0611-seed-data-methods}

\subsubsection{Seed Sampling Results}
\input{90x-appendices/0612-seed-data-results}

%% file: 90x-appendices/0611-seed-data-methods.tex
The seed data $B_i$
in tri-training is the labelled data that is used to train
the initial models.
This data is also included in the training data of the remaining
tri-training iterations in
our version of tri-training, see Algorithm~\ref{algo:tt}.
Each learner usually uses
a different sample of the original training data.

In initial experiments with the English side of the LinES Parallel Treebank
(\texttt{en\_lines}) as seed data,
we observed a degradation of performance
of the learners' models
when sampling the manually labelled
data with replacement -- as in vanilla tri-training~\cite{zhou-li-2005-tri} --
compared to models trained directly on the labelled data.
Neither combining three models in an ensemble
nor additional training data obtained through tri-training in
up to two tri-training
iterations compensated for the loss of performance.

The reason why vanilla tri-training uses sampling is to ensure variation
between the three learners.
Neural models, however, naturally vary due to random initialisation of
network weights, order of training data, stochastic kernels and
numerical effects when
intermediary results computed in parallel are combined in unpredictable
order.
We therefore tried using the original manually labelled training data in
all learners and relying on random initialisation to instill variation.
This removed the degradation of performance but as tri-training proceeded
performance stayed within 0.6 LAS points of the average LAS of
ensembles of three initial models.
We suspected that more variation is needed.
Therefore, we re-introduced sampling but modified it to ensure
that all manually labelled data is available to each learner.
We change the sampling to pick half of the data twice and the remaining half
three times, resulting in a sample size of 250\% of the original
data.\footnote{We confirmed that \udpf{} does not employ
     methods for handling unknown words based on applying
     a frequency threshold on the training data as
     oversampling may interfere with such methods.
}
With this sampling, tri-training performance clearly improved
in the \texttt{en\_lines} experiment and exceeded the range
of results due to random initialisation
and other sources of variation in neural models.

The results for our four development languages shown in Table~\ref{tab:seed-data-effect}
mostly confirm these findings.
Using 2.5 copies 
consistently gives the highest LAS, though
the improvements over vanilla tri-training, which
uses sampling only for the initial models and then
continues with the full labelled data, and a variant
using the full data from the start are small.

%% file: 90x-appendices/0612-seed-data-results.tex
\begin{table*}
\centering
\begin{tabular}{l|l|r|r|r|r}
\textbf{Lang.} & \textbf{Parser} & \textbf{W.R.} &
\textbf{Vanilla} & \textbf{100\%} & \textbf{250\%}\\
\hline
\multirow{2}{*}{En} & \textbf{udpf} & 75.3 & 77.2 & 77.9 & \textbf{78.0} \\
                   & \textbf{elmo} & 82.9 & 83.8 & 84.0 & \textbf{84.3} \\
\hline
\multirow{2}{*}{Hu} & \textbf{udpf} & 77.9 & 80.3 & 80.6 & \textbf{80.8} \\
                   & \textbf{elmo} & 84.1 & 85.3 & 85.5 & \textbf{85.5} \\
\hline
\multirow{2}{*}{Ug} & \textbf{udpf} & 66.4 & 67.9 & 68.1 & \textbf{68.3} \\
                   & \textbf{elmo} & 70.5 & 71.4 & 71.6 & \textbf{71.7} \\
\hline
\multirow{2}{*}{Vi} & \textbf{udpf} & 61.6 & 63.7 & 63.7 & \textbf{63.9} \\
                   & \textbf{elmo} & 66.3 & 67.5 & 67.6 & \textbf{67.6} \\
\hline
\multicolumn{2}{l|}{Average}& 73.13 & 74.63 & 74.87 & \textbf{75.01}\\
\hline
\end{tabular}
\caption{Effect of seed data sampling on tri-training performance (average development LAS over eight tri-training runs, selecting, for each run, the best tri-training iteration according to development LAS):
   W.R. is sampling with replacement,
   Vanilla uses sampling with replacement for the
   initial models and a full copy of the labelled
   data for $t>0$,
   100\% uses a full copy of the labelled data in
   all iterations,
   \ie the only source of variation is the random seed used in parser training, 250\% uses 2.5 copies
   of $L$ for each learner, providing additional variation due to the random selection
   of the last half of the data.
   }
\label{tab:seed-data-effect}
\end{table*}

%% file: 90x-appendices/0620-data-combination.tex
The tri-training parameter $d$ controls how much data from previous 
tri-training iterations is used in the current iteration.
We experiment with $d \in \{0, 0.5, 1\}$ as 
we expect that
          training a model on data obtained with different
          models, initialised with different seeds, may have
          similar benefits as using ensemble predictions,
          which \newcite{yu-etal-2020-ensemble}
          show to improve self-training.
          Furthermore, data combination
          may limit negative effects of
          an iteration with poorly performing
          models $h_i$.

\begin{table}
\centering
\begin{tabular}{l|l|r|r|r|r}
 & & \multicolumn{4}{c}{$d$} \\
\textbf{Lang.} & \textbf{Parser} & $0$ & $0.5$ & $0.71$ & $1$ \\
\hline
\multirow{2}{*}{En}     & \textbf{udpf} &         77.9  &         78.3  &         78.4  & \textbf{78.6} \\
                             & \textbf{elmo} &         84.1  &         84.3  &         84.4  & \textbf{84.9} \\
\hline
\multirow{2}{*}{Hu}   & \textbf{udpf} &         80.7  &         81.1  &         81.5  & \textbf{81.5} \\
                             & \textbf{elmo} &         85.5  &         85.7  & \textbf{85.8} &         85.8  \\
\hline
\multirow{2}{*}{Ug}      & \textbf{udpf} &         68.3  & \textbf{68.7} &         68.6  &         68.6  \\
                             & \textbf{elmo} &         71.6  &         71.8  &         71.8  & \textbf{72.0} \\
\hline
\multirow{2}{*}{Vi}  & \textbf{udpf} &         63.9  &         64.0  &         64.2  & \textbf{64.5} \\
                             & \textbf{elmo} &         67.3  &         67.7  & \textbf{67.9} &         67.8  \\
\hline
\end{tabular}
\caption{Effect of data combination across iterations on tri-training
    performance (average development LAS over multiple tri-training runs, selecting, for each run, the best tri-training iteration according to development LAS);
    two runs for each setting (language, parser, data combination method)
    with $A=80$k, $T=8$, seed data sampled as 250\% of labelled data
    and $o \in \{$True, False$\}$.
    }
\label{tab:data-combination-effect}
\end{table}

The results are shown in Table~\ref{tab:data-combination-effect}.
For all but the Uyghur parser \textbf{udpf}, \ie
without external word embeddings,
we found the best development results when predictions of
all tri-training iterations are combined.
The difference in LAS to the combination method that
exponentially reduces the size of data taken from
previous iterations ($d=0.5$) is small.

%% file: 90x-appendices/0630-augment-size.tex
\input{90x-appendices/0632-aug-table-horizontal}

The tri-training parameter $A$
controls how much unlabelled data is combined with labelled data during training.
Table~\ref{tab:augsize-effect}
presents results for augmentation
sizes $A$ from 5k to 160k tokens.\footnote{For
    comparison, the labelled data $L$ has about
    20k tokens in our experiments and the samples
    $B_i$ have about 50k tokens for our best
    seed data sample size 250\%.
}
We see good improvements for all development languages
except Vietnamese
as the size of the set of
automatically labelled data added in each tri-training
round increases.
For Vietnamese with parser \textbf{elmo},
the range of scores is small and there
is no consistent pattern.

%% file: 90x-appendices/0632-aug-table-horizontal.tex
\begin{table*}
\centering
\begin{tabular}{l|l||r|r|r||r|r|r||r|r}
 & & 
\multicolumn{3}{l||}{\textit{(i)} Small $A$} &
\multicolumn{3}{l||}{\textit{(ii)} Medium $A$} &
\multicolumn{2}{l}{\textit{(iii)} Big $A$} \\
\textbf{Language} & \textbf{Parser} & \textbf{5k} & \textbf{10k} & \textbf{20k} &
\textbf{20k} & \textbf{40k} & \textbf{80k} &
\textbf{80k} & \textbf{160k}  \\
\hline
\multirow{2}{*}{English} & \textbf{udpf} &
                         77.5 & 77.9 & \textbf{78.2} &
                         77.4 & 77.8 & \textbf{78.3} &
                         78.0 & \textbf{78.4} \\
                   & \textbf{elmo} & 
                         84.1 & 84.4 & \textbf{84.6} &
                         84.2 & 84.4 & \textbf{84.5} &
                         84.2 & \textbf{84.4} \\
\hline
\multirow{2}{*}{Hungarian} & \textbf{udpf} &
                         80.2 & 81.1 & \textbf{81.1} &
                         80.3 & 80.7 & \textbf{81.1} &
                         80.8 & \textbf{81.3} \\
                   & \textbf{elmo} &
                         85.3 & 85.7 & \textbf{85.7} &
                         85.3 & 85.4 & \textbf{85.7} &
                         85.6 & \textbf{85.8} \\
\hline
\multirow{2}{*}{Uyghur} & \textbf{udpf} &
                         67.5 & 67.9 & \textbf{68.5} &
                         67.6 & 67.9 & \textbf{68.4} &
                         68.2 & \textbf{68.4} \\
                   & \textbf{elmo} & 71.6 & 71.7 & \textbf{71.7} &
                         71.4 & 71.5 & \textbf{71.7} &
                         71.7 & \textbf{71.9} \\
\hline
\multirow{2}{*}{Vietnamese} & \textbf{udpf} &
                         63.6 & 63.7 & \textbf{64.2} &
                         63.4 & 63.8 & \textbf{64.2} &
                         63.9 & \textbf{64.0} \\
                   & \textbf{elmo} &
                         \textbf{67.8} & 67.7 & 67.7 &
                         67.5 & \textbf{67.6} & 67.6 &
                         67.6 & \textbf{68.0} \\
\hline
\end{tabular}
\caption{Effect of augmentation size $A$ on tri-training
    performance (average development LAS over multiple tri-training runs):
    \textit{(i)} two runs
        with $T=16$ and
        $d=1$,
    \textit{(ii)} three runs (Uygur) or four runs (other languages)
        with $T=8$ and
        $d=1$, 
    \textit{(iii)} six runs
        with $T=4$ and
        $d=0.5$.
    }
\label{tab:augsize-effect}
\end{table*}

%% file: 90x-appendices/0640-oversampling.tex
\begin{table}
\centering
\begin{tabular}{l|l|r|r|r}
\textbf{Language} & \textbf{Parser} & \textbf{No} & \textbf{Yes} & \textbf{$\Delta$}\\
\hline
\multirow{2}{*}{English} & \textbf{udpf} & 77.4 & \textbf{77.7} & 0.288 \\
 & \textbf{elmo} & 84.0 & \textbf{84.0} & 0.055 \\
\hline
\multirow{2}{*}{Hungarian} & \textbf{udpf} & 80.2 & \textbf{80.3} & 0.117 \\
 & \textbf{elmo} & 85.3 & \textbf{85.3} & 0.015 \\
\hline
\multirow{2}{*}{Uyghur} & \textbf{udpf} & 68.1 & \textbf{68.1} & 0.026 \\
 & \textbf{elmo} & \textbf{71.7} & 71.7 & -0.001 \\
\hline
\multirow{2}{*}{Vietnamese} & \textbf{udpf} & \textbf{63.8} & 63.8 & -0.022 \\
 & \textbf{elmo} & \textbf{67.7} & 67.7 & -0.048 \\
\hline
\end{tabular}
\caption{Effect of oversampling the labelled data on tri-training
    performance (average development LAS over multiple tri-training runs)
    }
\label{tab:oversampling-effect}
\end{table}

Table~\ref{tab:oversampling-effect}
compares average LAS with and without oversampling
of the manually labelled data $B_i$ to match the size of the automatically labelled data $R$.
The results suggest that the
effects is negligible and since oversampling slows down training we
carry out the main experiment without
oversampling.\footnote{Preliminary
    results for English
    with oversampling the manually labelled data
    three times in all iterations, including the seed models
    (Table~\ref{tab:seed-data-effect}), however,
    show a positive effect of oversampling.
    Maybe oversampling is more important in early iterations where the
    amount of automatically labelled data is relatively small.
    Future work should investigate the effect of oversampling
    further.
}

%% file: 90x-appendices/0760-error-analysis.tex
Table~\ref{table:LASbyLabel} shows the most frequent LAS improvements by dependency label.
\begin{table*}
\begin{tabular}{|l|l|l|l|l|}
\textbf{Parsers} & \textbf{English} & \textbf{Hungarian} & \textbf{Uyghur} & \textbf{Vietnamese} \\\hline
\multirow{3}{*}{\textcolor{ttblue}{base udpf} -$>$ \textcolor{ttred}{tri udpf}} 
& 
acl,fixed &
nummod,csubj
& parataxis,mark & compound,mark
\\
&
compound,xcomp & cop,nsubj
& obj,nummod &
xcomp,cop
\\ 
&
parataxis,iobj & case,advmod
& aux,nsubj & 
case,csubj
\\
  \hline
\hline
\multirow{3}{*}{\textcolor{ttblue}{base udpf} -$>$ \textcolor{ttorange}{base elmo}} 
& 
\textbf{flat},discourse & 
cop,advcl  &
appos,conj & 
amod,ccomp 
\\
& 
fixed,parataxis & 
nsubj,acl &
parataxis,\textbf{flat} & 
mark,obj
\\
& 
acl,appos & 
\textbf{flat},ccomp &
fixed,nummod & 
discourse,compound
\\ 
  \hline
  \hline
\multirow{3}{*}{\textcolor{ttorange}{base elmo} -$>$ \textcolor{ttgreen}{tri elmo}}
& 
fixed,ccomp & 
nummod,csubj
 & discourse,appos & 
csubj,mark 
\\
& 
advcl,flat & 
advcl,appos
 & cop,compound & 
parataxis,compound
\\
& 
obl,punct & 
acl,parataxis
 & ccomp,obl & 
amod,cop
\\
  \hline
\end{tabular}
\caption{Top 6 largest LAS improvements by dependency type. Those shared by at least 3 languages are highlighted in bold. Labels with fewer than 20 occurrences are excluded.}\label{table:LASbyLabel}
\end{table*}

%% file: 90x-appendices/0730-learning-rate-scehd.tex
\input{90x-appendices/0352-b-learning-rate}

Table~\ref{table:mtb:schedules} compares the learning rate schedule
we use with UDPipe-Future and its default schedule,

%% file: 90x-appendices/0352-b-learning-rate.tex
\begin{table*}
\centering
\begin{tabular}{|l|l|rrrrrr|}
\textbf{} &
\textbf{Total} & \multicolumn{6}{|c|}{\textbf{Learning Rate}} \\
\textbf{Setting} &
\textbf{Epochs} & \textbf{0.001} & \textbf{0.0006} & \textbf{0.0004} & \textbf{0.0003} & \textbf{0.0002} & \textbf{0.0001} \\
\hline
Parser default &
60 & 40 &   0 &   0 & 0 & 0   & 20 \\
\hline
In this work &
60 & 30 &   5 &   5 & 5 & 5   & 10 \\
\hline
\end{tabular}
\caption{Learning rate schedule used in the experiments and the parser's default learning rate: number of epochs at each learning rate}
\label{table:mtb:schedules}
\end{table*}

%% file: 90x-appendices/0380-model-selection.tex
We select the tri-training iteration with the best
ensemble performance according to development LAS.
We do not use \newcite{zhou-li-2005-tri}'s
stopping criterion that is based on conditionally
updating the learners' models
in line~\ref{algo:tt:update} of Algorithm~\ref{algo:tt}
for the following reasons:
\begin{itemize}

    \item The model update condition is designed for
          binary classification tasks.
          It is not clear how the condition would have to be
          updated for joint prediction of
          dependency trees, lemmata and multiple tags.

    \item The model update condition uses the
          training data to estimate label noise.
          We do not expect such estimates to be
          useful for neural models that tend
          to considerably overfit the training data.

    \item The model update condition rejects models
          trained on an amount of automatically
          labelled data that is too small
          to avoid harm from label noise
          under certain assumptions.
          In \newcite{zhou-li-2005-tri}'s experiments, the
          size of the unlabelled data is quite
          small.\footnote{
              \newcite{zhou-li-2005-tri}'s largest dataset has
              only 3772 items. The
              size of the unlabelled data never exceeds 
              $3772 \times 0.75 \times 0.80 \approx 2263$ items.
          }
          In contrast, we can avail of orders of magnitude
          more unlabelled data.
          Hence, we do not expect the issue of insufficient
          data to arise.

    \item The inherent performance variation of
          neural models, \eg due to random initialisation,
          can trigger \newcite{zhou-li-2005-tri}'s stopping
          criterion too early as it requires the error rate
          to drop in each iteration.
          When tri-training is run long enough for the
          improvements due to the additional training data
          to be smaller than the performance variation due
          to randomness in model training,
          we expect that patience is needed to bridge a
          temporary degradation.\footnote{We
              do find variation in performance
              and performance recovery after a few
              iterations looking at a sample of tri-training runs.
              Future work can analyse our data for the trade-off
              between the lengths of patience and compute costs
              (or spending the same compute budget on more runs
              with lower patience each).
          }

    \item Furthermore, tri-training can reduce performance
          if wrong decisions are amplified.

\end{itemize}

%% file: 90x-appendices/0750-bert-layers.tex
We experiment with using different BERT layers and pooling functions for combining BERT's
subword vectors to token vectors.
We explore 45 settings for each language, nine choices of layers
(individual layers and average of layers, excluding bottom layers)
and five choices of token pooling functions.
For each language, we choose three different settings, one for each
learner in tri-training, 
starting with the top-performing setting, eliminating all settings with
the same choice of layers or the same choice of token pooling and then
repeating the process for the next learner.

\begin{table*}
\centering
\begin{tabular}{l|r|r|r|r|r|r|r|r|r|r|}
\multicolumn{11}{l}{} \\
\multicolumn{11}{l}{English} \\
\hline
Pooling \textbackslash{} Layer & 08 & 09 & 10 & 11 & 12 & A4 & A4E & A5 & A5E & Average \\
\hline
Avg   & \textbf{85.5} & 85.4 & 85.3 & 84.7 & 83.0 & 85.1 & 85.2 & 85.4 & 85.4 & 85.0 \\
First & 85.3 & 85.3 & 85.1 & 84.6 & 83.0 & 84.9 & 85.1 & 85.3 & 85.3 & 84.9 \\
Last  & 85.4 & 85.4 & 85.2 & 84.6 & 83.0 & 85.0 & 85.1 & 85.3 & \textbf{85.3} & 84.9 \\
Max   & 85.3 & 85.4 & 85.2 & 84.6 & 83.0 & 85.0 & 85.2 & 85.2 & 85.2 & 84.9 \\
Z50   & 85.4 & \textbf{85.4} & 85.2 & 84.7 & 83.0 & 85.0 & 85.2 & 85.3 & 85.4 & 85.0 \\
\hline
Average & 85.4 & 85.4 & 85.2 & 84.7 & 83.0 & 85.0 & 85.2 & 85.3 & 85.3 & -- \\
\hline
\multicolumn{11}{l}{} \\
\multicolumn{11}{l}{Hungarian} \\
\hline
Pooling  \textbackslash{} Layer & 08 & 09 & 10 & 11 & 12 & A4 & A4E & A5 & A5E & Average \\
\hline
Avg   & 85.0 & 85.2 & 85.1 & 84.6 & 83.4 & 85.0 & 84.9 & 85.0 & \textbf{85.1} & 84.8 \\
First & 84.9 & 84.8 & 84.7 & 84.4 & 82.9 & 84.6 & 84.6 & 84.5 & 84.6 & 84.4 \\
Last  & 85.3 & \textbf{85.5} & 85.3 & 85.0 & 83.6 & 85.2 & 85.3 & 85.4 & 85.3 & 85.1 \\
Max   & \textbf{84.9} & 85.2 & 84.9 & 84.5 & 83.1 & 84.7 & 84.7 & 84.8 & 84.9 & 84.6 \\
Z50   & 84.9 & 85.1 & 84.9 & 84.5 & 83.3 & 84.9 & 84.8 & 84.8 & 84.8 & 84.6 \\
\hline
Average & 85.0 & 85.2 & 85.0 & 84.6 & 83.3 & 84.9 & 84.9 & 84.9 & 84.9 & -- \\
\hline
\multicolumn{11}{l}{} \\
\multicolumn{11}{l}{Uyghur} \\
\hline
Pooling  \textbackslash{} Layer & 08 & 09 & 10 & 11 & 12 & A4 & A4E & A5 & A5E & Average \\
\hline
Avg   & 66.8 & 67.0 & 66.8 & 66.8 & 67.2 & 66.8 & 67.0 & \textbf{67.2} & 66.9 & 66.9 \\
First & 66.9 & 66.8 & 66.9 & 66.9 & 67.1 & 66.9 & 67.0 & 67.0 & 67.0 & 66.9 \\
Last  & 66.5 & 66.6 & 66.8 & 66.7 & 66.8 & 66.8 & 66.6 & 66.8 & 66.8 & 66.7 \\
Max   & \textbf{67.0} & 66.8 & 66.7 & 66.6 & 67.0 & 66.9 & 66.9 & 66.9 & 66.8 & 66.9 \\
Z50   & 67.0 & 67.0 & 66.9 & 66.8 & \textbf{67.2} & 66.9 & 66.8 & 67.0 & 66.9 & 66.9 \\
\hline
Average & 66.8 & 66.9 & 66.8 & 66.8 & 67.1 & 66.9 & 66.9 & 67.0 & 66.9 & -- \\
\hline
\multicolumn{11}{l}{} \\
\multicolumn{11}{l}{Vietnamese} \\
\hline
Pooling  \textbackslash{} Layer & 08 & 09 & 10 & 11 & 12 & A4 & A4E & A5 & A5E & Average \\
\hline
Avg   & 68.5 & \textbf{68.6} & 68.4 & 68.3 & 67.4 & 68.5 & 68.3 & 68.5 & 68.4 & 68.3 \\
First & 68.7 & 68.3 & 68.4 & 68.1 & 67.6 & 68.2 & 68.3 & \textbf{68.7} & 68.4 & 68.3 \\
Last  & \textbf{68.7} & 68.4 & 68.5 & 68.2 & 67.2 & 68.4 & 68.3 & 68.6 & 68.5 & 68.3 \\
Max   & 68.5 & 68.3 & 68.3 & 68.1 & 67.5 & 68.4 & 68.2 & 68.3 & 68.4 & 68.2 \\
Z50   & 68.5 & 68.5 & 68.3 & 67.9 & 67.6 & 68.3 & 68.3 & 68.4 & 68.5 & 68.3 \\
\hline
Average & 68.6 & 68.4 & 68.4 & 68.1 & 67.5 & 68.4 & 68.3 & 68.5 & 68.5 & -- \\
\hline
\multicolumn{11}{l}{} \\
\multicolumn{11}{l}{Average over all languages} \\
\hline
Pooling  \textbackslash{} Layer & 08 & 09 & 10 & 11 & 12 & A4 & A4E & A5 & A5E & Average \\
\hline
Avg   & 76.5 & 76.6 & 76.4 & 76.1 & 75.3 & 76.3 & 76.4 & 76.5 & 76.4 & 76.3 \\
First & 76.4 & 76.3 & 76.3 & 76.0 & 75.1 & 76.1 & 76.2 & 76.4 & 76.3 & 76.1 \\
Last  & 76.5 & 76.5 & 76.4 & 76.1 & 75.2 & 76.4 & 76.3 & 76.5 & 76.5 & 76.3 \\
Max   & 76.4 & 76.5 & 76.3 & 76.0 & 75.1 & 76.3 & 76.3 & 76.3 & 76.3 & 76.2 \\
Z50   & 76.4 & 76.5 & 76.3 & 76.0 & 75.3 & 76.3 & 76.3 & 76.4 & 76.4 & 76.2 \\
\hline
Average & 76.4 & 76.5 & 76.3 & 76.0 & 75.2 & 76.3 & 76.3 & 76.4 & 76.4 & -- \\
\hline
\end{tabular}
\caption{Development set LAS for training UDPipe-Future with word embeddings taken from
different BERT layers. Each inner cell shows the average over 25 runs. Pooling methods
are average, first, last, maximum and weighted average with binonimal distribition with
$p=0.5$ (Z50).
Layer A4 (A5) stands for using the average of the top 4 (5) layers.
The E suffix means that the 768-dimensional BERT\_BASE vectors are expanded to 1024
components so that all (A4E) or most (A5E) final components can be the average
of fewer input components.
The three settings selected for the three learners of each language are shown in \textbf{bold}.
}
\label{tab:bert-layers}
\end{table*}

Table~\ref{tab:bert-layers} shows the results of this experiment.
Our observations confirm that middle layers typically perform best \cite{rogers-etal-2020-primer}.
Uyghur, for which multilingual BERT does not perform well with UDPipe-Future,
has a different pattern showing no large differences and a preference for
the top layer that is less informative for the other languages.
Different languages seem to prefer different pooling functions for combining
vectors of subword units to vectors for UD tokens.
Table~\ref{tab:bert-layers} shows our choices for each development language
in bold.